\pdfoutput=1
\documentclass[11pt]{article}
\usepackage{acl}
\usepackage{times}
\usepackage{latexsym}
\usepackage[T1]{fontenc}
\usepackage[utf8]{inputenc}
\usepackage{braket}
\usepackage{microtype}
\usepackage{inconsolata}
\usepackage{amsmath}
\usepackage{amssymb}
\usepackage{amsfonts}
\usepackage{tabularx}
\usepackage{textcmds}
\usepackage{url}
\usepackage{bm}
\usepackage{tikz}
\usepackage{makecell}
\usepackage[export]{adjustbox}
\usepackage{tabstackengine}
\usepackage{booktabs}
\usepackage{multirow}
\usepackage{subfig}
\usepackage[normalem]{ulem}

\usepackage{soul}

\newcommand{\m}[1]{\mathbf{#1}}

\newcommand{\tabincell}[2]{\begin{tabular}{@{}#1@{}}#2\end{tabular}}

\newcommand{\TODO}[1]{}
\title{Visually Guided Generative Text-Layout Pre-training \\ for Document Intelligence}
\author{Zhiming Mao$^{1,2}$, Haoli Bai$^{3}$, Lu Hou$^{3}$, \\
\textbf{Jiansheng Wei$^{3}$, Xin Jiang$^{3}$, Qun Liu$^{3}$, Kam-Fai Wong$^{1,2}$} \\
$^1$The Chinese University of Hong Kong, Hong Kong, China \\
$^2$MoE Key Laboratory of High Confidence Software Technologies, China \\
$^3$Noah’s Ark Lab, Huawei Technologies \\
\{zmmao, kfwong\}@se.cuhk.edu.hk \\
\{baihaoli, houlu3, weijiansheng, jiang.xin, qun.liu\}@huawei.com}
\begin{document}
\maketitle
\begin{abstract}
Prior study shows that pre-training techniques can boost the performance of visual document understanding~(VDU), which typically requires models to gain abilities to perceive and reason both document texts and layouts (e.g., locations of texts and table-cells). To this end, we propose visually guided generative text-layout pre-training, named ViTLP. Given a document image, the model optimizes hierarchical language and layout modeling objectives to generate the interleaved text and layout sequence. In addition, to address the limitation of processing long documents by Transformers, we introduce a straightforward yet effective multi-segment generative pre-training scheme, facilitating ViTLP to process word-intensive documents of any length. ViTLP can function as a native OCR model to localize and recognize texts of document images. Besides, ViTLP can be effectively applied to various downstream VDU tasks. Extensive experiments show that ViTLP achieves competitive performance over existing baselines on benchmark VDU tasks, including information extraction, document classification, and document question answering\footnote{Code and checkpoint will be available at \href{https://github.com/Veason-silverbullet/ViTLP}{https://github .com/Veason-silverbullet/ViTLP}.}.
\end{abstract}
\section{Introduction}
Processing and reasoning document images with dense texts (e.g., scanned PDF files, digital forms, and spreadsheets) is a persistent yet challenging task for the research community and industry \citep{Chargrid, DocIE, structurallm}. Advances in multimodal pre-training substantially improve the performance of visual document understanding (VDU) \citep{LayoutLM-v1, LayoutLM-v2, UniDoc, DocFormer, LILT}. These pre-training methods typically take multimodal inputs of given document images including \romannumeral 1) visual features, \romannumeral 2) pre-processed OCR texts, and \romannumeral 3) spatial layouts of document elements (e.g., $2$D coordinates of texts and table-cells). Among these inputs, spatial layout information plays an essential role in connecting visual and textual features, as well as developing thorough reasoning of document structures \citep{WebSRC, FormNet}.

\begin{figure}
    \centering
    \vskip -2.0988mm
    \includegraphics[width=1\linewidth]{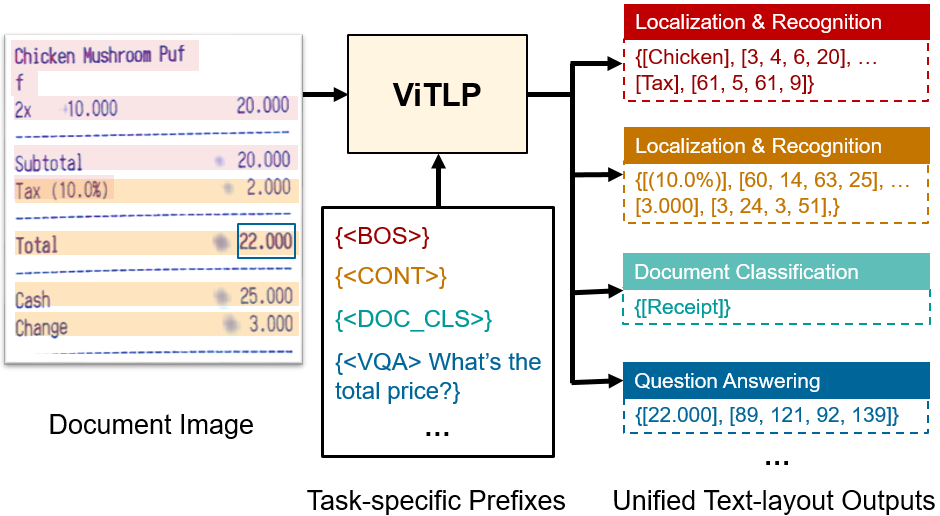}
    \vskip -0.4mm
    \caption{An overview workflow of the proposed ViTLP. Given a document image as input, ViTLP can generate sequences of text and layout (i.e., word bounding boxes) for various VDU tasks with task-specific prefixes.}
    \vskip -0.4mm
    \label{fig:intro}
\end{figure}

Though effective, the performance of most existing VDU approaches relies heavily on the OCR pipelines, because the pre-processed OCR texts and corresponding $2$D coordinates are used as intermediate inputs to pre-trained VDU models. The external OCR pipelines may produce incorrect or incomplete recognition results, which cannot be jointly optimized by the gradient back from VDU models. Another research line \citep{DONUT, Pix2Struct} explores pre-training VDU models solely based on image inputs. Despite no OCR errors introduced, these methods focus on understanding texts from raw document images but neglect layout information modeling. Since the spatial information contained in layout locations is not exploited, it may hinder the models from understanding complex document structures, especially for documents containing nested paragraphs, forms, and tables.

In this work, we propose \textbf{Vi}sually guided generative \textbf{T}ext-\textbf{L}ayout \textbf{P}re-training~(ViTLP) to jointly model text and layout information from document images. As shown in Figure~\ref{fig:intro}, ViTLP can localize, recognize, and understand visual document texts given the input document image and task prefixes. To achieve this goal, ViTLP is pre-trained to generate \textit{unified text-layout sequences} from document images. Since natively generating text and layout tokens in a flattened sequence is \textit{token-inefficient} (see Sec. \ref{sec:problem}), we introduce hierarchical generation modules to achieve both effective and efficient text-layout sequence generation. To the best of our knowledge, ViTLP is the first attempt to learn OCR (i.e., text localization and recognition) and VDU (i.e., document understanding) abilities in a unified generative text-layout pre-training framework.

Besides, ViTLP is designed to handle long documents with intensive texts. Long document processing is ubiquitous in real-world scenarios. However, existing pre-trained models are constrained to certain token limits of input sequences. For instance, LayoutLMv$2$ \citep{LayoutLM-v2} accepts the maximum inputs of $512$ word tokens using a BERT-structure encoder. In both pre-training and fine-tuning, the exceeded text tokens are truncated, leading to incomplete document information modeling. To tackle this issue, we introduce a \textbf{\mbox{multi-segment} pre-training scheme} which divides the target text-layout sequence into consecutive segments to perform generative pre-training. Given that the full document information is already encoded in visual representations, ViTLP takes the suffix tokens from previous segments as prefix prompts to generate the next-segment tokens. This multi-segment pre-training scheme further enables ViTLP to process documents of arbitrary length in fine-tuning. Notably, our multi-segment generation scheme retains the intact transformer architecture. Thus, it is more feasible than other long-document modeling workarounds, e.g., \textit{sparse attention} \citep{Longformer} and \textit{memory modules} \citep{RMT}, which need to modify the Transformer architecture and may affect the capacity of pre-trained models.

We evaluate ViTLP on a variety of OCR and VDU tasks. Experiment results demonstrate that ViTLP can achieve superior overall performance on both OCR and VDU tasks. For instance, ViTLP achieves the $95.59\%$ F$1$ score on CORD information extraction and $95.36\%$ accuracy on RVL-CDIP document classification, both of which outperform most previous approaches. Notably, ViTLP can intrinsically generate $2$D layout locations for visual grounding, which helps in certain generative VDU tasks (e.g., visual document question answering) to be more interpretable and reliable to humans.

\section{Approach}
\label{sec:method}
\begin{figure*}[t]
\centering
\includegraphics[width=160mm]{Figures/Model.pdf}
\captionsetup{font=10pt}
\caption{Overview of the ViTLP architecture. ViTLP is a generative pre-training model that performs autoregressive text-layout modeling conditioned on visual document inputs. ViTLP adopts hierarchical decoder heads to generate target text-layout sequences in a \textit{global-to-local} manner. The segment mode tokens $\in\{ \texttt{[BOS]}, \texttt{[CONT]}\}$ prompt the beginning and continuous modes of generation, respectively.}
\label{Figures:model}
\end{figure*}
\subsection{Problem Formulation}\label{sec:problem}
We study multimodal pre-training for visual document modeling. As widely studied \citep{LayoutLM-v1, LayoutLM-v2, DocFormer, SelfDoc, TILT, LILT, LayoutLM-v3, MGDoc}, document images $\m V$, texts $\m T$, and layouts $\m L$ are three fundamental modalities for visual document modeling.

\paragraph{Unified Text-Layout Generation.} We cast the pre-training objective on visual documents as text-layout sequence (i.e., $\{\m T; \m L\}$) generation conditioned on document images $\m V$. The document texts $\m T$ are represented as word-token sequences. The layouts $\m L$, following prior studies \citep{LayoutLM-v1, LayoutLM-v2}, can be represented by \textit{location bounding boxes} of words. Instead of generating two separate sequences of $\m T$ and $\m L$, ViTLP generates the texts with corresponding layout locations in a sequence of interleaved text-layout tokens, which facilitates compact multimodal interaction between texts and layouts. For the $i$-th word of a document, its text-layout tokens $\{\m T; \m L\}_i$ are represented as
\begin{align}\label{eq:seq}
    \{\m T; \m L\}_i = \big\{ \{\boldsymbol w\}_i, \{z_{x_1}, z_{y_1}, z_{x_2}, z_{y_2} \}_i \big\},
\end{align}
where $\{\boldsymbol w\}_i$ denotes the BPE tokens \cite{GPT2} of the $i$-th word, $\{z_{x_1}, z_{y_1}, z_{x_2}, z_{y_2} \}_i \in \mathbb{Z}_{+}^4$ are the corresponding left-top and right-bottom bounding box coordinates. Given a document with $N$ words, the objective is to maximize the likelihood function $\log p(\m T; \m L|\m V)$ which can be decomposed as autoregressive text and layout modeling:
\vskip-4.33925629mm
\begin{align}\label{eq:obj}
    \log p(\m T; \m L|\m V)& = \sum_{i=1}^{N} \big(\underbrace{\log p(\m T_i | \m T_{< i}, \m L_{< i}, \m V)}_{\textrm{Text-modeling}} \nonumber \\
    & + \underbrace{\log p(\m L_i | \m T_{\leq i}, \m L_{< i}, \m V)}_{\textrm{Layout-modeling}}\big).
\end{align}

Note that Eq.~\eqref{eq:obj} shares similar ideas with \citet{Pix2seq}, where word and bounding box generation can be formulated as language modeling on a unified text-layout sequence. However, it is in fact nontrivial to generate sequences as in Eq.~\eqref{eq:seq}, because real-world documents commonly contain intensive texts, generating \ul{each word followed by four coordinate tokens} in a long flattened sequence is especially \textbf{token-inefficient}. This would bring prohibitive computational and space overhead\footnote{Recall that both the computational and space complexities of Transformers are quadratic $\mathcal{O}(L^2)$ in sequence length $L$.} to the Transformer-based text-layout decoder.

\subsection{Model Architecture}
\label{sec:model_arch}
The architecture of ViTLP is shown in Figure~\ref{Figures:model}. ViTLP employs an encoder-decoder framework to encode document images $\m V$ and generate target text-layout sequences $\{\m T; \m L\}$. Specifically, given an input document image $\m V$, ViTLP employs a vision transformer (ViT) \citep{VIT} to learn visual representations $\m H^{V} \in \mathbb{R}^{|V|\times d}$, where $|V|$ is the ViT patch number and $d$ is the hidden size. The decoder receives the visual representations $\m H^V$ and generates the unified text-layout sequence $\{\m T; \m L\}$. To address the \textit{token-inefficiency} issue discussed in Sec.~\ref{sec:problem}, we design the \textit{global-to-local} text-layout generation process as follows.

\subsubsection{Global Text-Layout Modeling}\label{global_text_layout_modeling}
Instead of directly generating the text-layout sequence as in Eq.~\eqref{eq:seq}, we first replace the bounding box coordinates $\{z_{x_1}, z_{y_1}, z_{x_2}, z_{y_2}\}$ with a generic layout location token $\Hat{w}=\texttt{[LOC]}$. This integrates the mixed text-layout sequence $\{\m T; \m L\}$ to unified language modeling. Given the original vocabulary $\mathcal{V}$, the \textbf{global text-layout sequence} $\Hat{\m T}$ derives from the augmented vocabulary $\mathcal{\Hat{V}}=\mathcal{V} \cup \texttt{[LOC]}$. The layout token embeddings $\mathrm{E}_{\texttt{[LOC]}}$ are computed as
\begin{equation*}
\mathrm{E}_{\texttt{[LOC]}}=\big[\mathrm{E}_{x}(z_{x_1}), \mathrm{E}_{y}(z_{y_1}), \mathrm{E}_{x}(z_{x_2}), \mathrm{E}_{y}(z_{y_2})\big],
\end{equation*}
where $\mathrm{E}_{x}(\cdot)\in\mathbb{R}^{\frac{d}{4}}$ and $\mathrm{E}_{y}(\cdot)\in\mathbb{R}^{\frac{d}{4}}$ denote the x- and y-axis spatial embeddings. Besides, the word tokens are embedded by $\mathrm{E}_{w}(\cdot)\in\mathbb{R}^{d}$. Given a document of $N$ words and the corresponding bounding boxes, the text-layout input embeddings are represented as $\m H^{TL}=\{\mathrm{E}_{w}, \mathrm{E}_{\texttt{[LOC]}}\}\in\mathbb{R}^{|\Hat{\m T}| \times d}$.

The ViTLP text-layout decoder performs multimodal interaction among \textit{visual}, \textit{textual}, and \textit{layout} information via the Transformer cross-attention
\begin{equation}\label{eq1}
    \m H^{VTL} = \mathrm{Transformer\mbox{-}Decoder}(\m H^{V}, \m H^{TL}).
\nonumber
\end{equation}
For the $i$-th target token $\Hat{\m T}_i$, the multimodal decoder output $\m H_i^{VTL}$ is fed to a linear language modeling (LM) head with the softmax function to compute the conditional generative probability
\begin{equation}
p(\Hat{\m T}_{i}|\Hat{\m T}_{<i}, \m V)=\textrm{Softmax}\big(\textrm{Linear}(\m H^{VTL}_{i})\big). \nonumber
\end{equation}
With the generic layout token \texttt{[LOC]} incorporated, the text-modeling term in Eq.~\eqref{eq:obj} is expressed as
\begin{equation}
\label{eq:global_loss}
\mathcal{L}_{\textrm{global-text}}=-\frac{1}{|\Hat{\m T}|}\sum\limits_{i=1}^{|\Hat{\m T}|} \log p(\Hat{\m T}_{i}|\Hat{\m T}_{<i}, \m V).
\end{equation}

\subsubsection{Local Layout Modeling}\label{local_layout_modeling}
Local layout modeling aims to generate specific layout locations for each generic layout token \texttt{[LOC]}. To capture the spatial relation among coordinates, we employ a lightweight sequential MLP layout head (see details in Appendix \ref{Appendix_RNN_layout_head}) to decode the \ul{short sequence of four layout coordinate tokens} from the last hidden state of \texttt{[LOC]}. For notation simplicity, we denote $\{\m L_{i,j}\}_{j=1}^4 = \{z_{x_1}, z_{y_1}, z_{x_2}, z_{y_2}\}_i$ as the corresponding layout coordinates of the \texttt{[LOC]} token at the $i$-th position, and its generative probability is modeled as
\begin{equation}
    p(\m L_{i,j}|\Hat{\m T}_{\leq i}, \m L_{i,<j}, \m V)=\textrm{Softmax}\big(\mathrm{MLP}(\m H_{i,<j})\big), \nonumber
\end{equation}
where $\m H_{i,0} = \m H_i^{VTL}$ is selected from the learned multimodal representations where $\Hat{\m T}_{i}=\texttt{[LOC]}$. Here, we denote the index set of \texttt{[LOC]} tokens as $\mathcal{S}_L=\big\{i: \Hat{\m T}_{i}=\texttt{[LOC]}|\,i=1,2,...,|\Hat{\m T}|\big\}$. The layout-modeling term in Eq.~\eqref{eq:obj} is expressed as
\begin{align}\label{eq:local_loss}
& \mathcal{L}_{\textrm{local-layout}} = -\sum\log p(\m L_i | \Hat{\m T}_{\leq i}, \m L_{< i}, \m V) \\
& = -\frac{1}{4|\mathcal{S}_L|}\sum_{i\in\mathcal{S}_L}\sum_{j=1}^{4} \log p(\m L_{i,j}|\Hat{\m T}_{\leq i}, \m L_{i,<j}, \m V). \nonumber 
\end{align}

In summary, with the global and local text-layout modeling in a hierarchy, the original pre-training objective in Eq.~\eqref{eq:obj} evolves to 
\begin{equation}\label{eq:final_obj}
    \mathcal{L}  = \mathcal{L}_{\textrm{global-text}} + \mathcal{L}_{\textrm{local-layout}}.
\end{equation}

The \textit{global-to-local} generation process aims to be effective and efficient for text-layout \mbox{modeling.} On effectiveness, the interleaved text-layout sequence modeling enables compact interaction between text and layout inputs, which can effectively fuse the information of text and layout modalities. On efficiency, suppose that the average BPE tokens of a document word are $|w|$, and the \textit{compression ratio} of the text-layout sequence is $\frac{|w|+1}{|w|+4}$, i.e., four coordinate tokens are compressed to one. In our experiment datasets, the \textit{compression ratio} is $0.48$.

\subsection{Multi-segment Pre-training Scheme}
\label{sec:pre-training_scheme}
Documents are usually intensive in text and layout, and it would be computationally intractable to fit the entire sequence into a generative model. To process documents with arbitrary length, we propose a multi-segment pre-training scheme that divides the long sequence into multiple segments for generation. Since a document image already contains all necessary information of the text and layout, long document modeling is feasible based on the \textit{visual representations} and \textit{localized generation-context}.

Given the maximum sequence length of the decoder as $M$, we first divide the text-layout sequence into $K$ segmented sequences $\{\m S_{i}\}_{i=1}^{K}$. The beginning segment $\m S_{1}$ contains $M$ tokens to be generated, and the continuous segment $\m S_{i>1}$ contains $\alpha_{p} \cdot M$ prefix tokens and $(1-\alpha_{p}) \cdot M$ tokens to be generated. Here, $\alpha_{p}$ is the pre-defined prefix ratio. The overall generation process comprises beginning and continuous modes.

\paragraph{Beginning Generation Mode.} In this mode, we prepend a special mode token \texttt{[BOS]} to the beginning sequence $\m S_{1}$. The model then follows the objective in Eq.~\eqref{eq:final_obj} to generate the first $M$ tokens.

\paragraph{Continuous Generation Mode.} For the continuous segments $\m S_{i>1}$, we prepend a special mode token \texttt{[CONT]} to the input sequence. $|P|=\alpha_{p} \cdot M$ prefix tokens are prepended to the input sequence. These $|P|$ \textbf{prefix tokens} of segmented sequence $\m S_{i}$ come from the $|P|$ \textbf{suffix tokens} of the previous segmented sequence $\m S_{i-1}$. The prefix tokens serve as a prompt of \textit{localized generation-context}\footnote{The historical context contains the generated coordinate tokens from the previous segment, which serves as an informatively complete prompting signal for next-segment generation.} which guides the decoder to generate subsequent tokens from arbitrary locations of a document. The special token \texttt{[EOS]} is appended to the last segmented sequence $\m S_{K}$ to signal the end of generation.

\paragraph{Segmentation in Pre-training and Fine-tuning.} In pre-training, the segmented sequences of a long document are randomly scattered into different data batches. In this way, ViTLP learns to model the complete textual and layout information of a document, conditioned on different prefix history-token contexts. In fine-tuning (and inference), ViTLP can also apply the multi-segment scheme to process those long text-layout sequences, which is consistent with the pre-training phase. For instance, OCR and sequence labeling on long document texts can be processed segment by segment.

\begin{table*}[t]
\centering
\fontsize{7.7}{7.7}\selectfont
\begin{tabular}{l|cccccc}
\toprule
\multirow{2}{*}{Approach} & \multicolumn{2}{c}{OCR Tasks} & \multicolumn{4}{c}{VDU Tasks} \\
 & Text Local. & Text Recog. & Info. Extraction & Doc. Classification & Document VQA & VQA Grounding  \\
\midrule
OCR Pipelines & \checkmark & \checkmark & & \\
Discriminative VDU Models & & & \checkmark & \checkmark & \checkmark & \\
Generative VDU Models &  &  & \checkmark & \checkmark & \checkmark & \\
ViTLP & \checkmark & \checkmark & \checkmark & \checkmark & \checkmark & \checkmark\\
\bottomrule
\end{tabular}
\caption{The comprehensive capabilities of ViTLP and its comparison with the associated baselines on each task.}
\label{tables:comparison}
\end{table*}
\subsection{Applications of ViTLP}
\label{sec:applications}
\subsubsection{OCR Text Localization and Recognition}
Text localization and recognition are two fundamental functions of OCR engines \citep{TrOCR}. As ViTLP is pre-trained to generate text and layout (i.e., $2$D bounding boxes) sequences from document images, it can intrinsically perform text localization and recognition by generating a unified OCR sequence of texts and bounding boxes. ViTLP can function as a word-level OCR model.

\subsubsection{Downstream VDU Tasks}\label{VDU_fine-tuning}
\paragraph{Information Extraction.} The information extraction task is formulated as sequence labeling on the target texts given document image input. Following BART \citep{BART}, we feed ViTLP decoder's final hidden states of a target word (with layout coordinate inputs) to a linear classifier which outputs the token-level semantic label.

\paragraph{Document Classification.} Given an input document image to the encoder, we feed a task prefix token \texttt{[DOC\_CLS]} as input to the decoder to output the document classification label.

\paragraph{Document Visual Question Answering.} Unlike discriminative VDU models that perform extractive QA on pre-processed OCR results, ViTLP directly generates answers given a task prefix token \texttt{[VQA]} followed by the question. It is noteworthy that ViTLP can intrinsically generate interpretable grounding \textbf{regions of interest (ROI)}, i.e., layout coordinates of answers, to verify the generation.

\section{Experiments}
\subsection{Experiment Setup}
\paragraph{Implementation Details.} We implement ViTLP with a $12$-layer ViT \citep{VIT} image encoder and a $6$-layer text-layout decoder. The Transformer hidden size is $d=768$ with $12$ attention heads. In pre-training, the input image height and width are $1920\mathrm{\times}1600$ with the $32\mathrm{\times}32$ ViT patch size, and the decoder segmented sequence length is $M=1024$. Following LayoutLMv$2$ \citep{LayoutLM-v2}, the layout location coordinates are normalized into discrete bins of $[0, 1000]$, resulting that the vocabulary size of the layout head is $1001$. The multi-segment prefix ratio is set as $\alpha_{p}=0.25$. We use the AdamW optimizer \citep{AdamW} to train ViTLP in $250$K steps, with the batch size of $384$ and initial learning rate of $2e$-$4$ with cosine decay. More implementation details are provided in Appendix \ref{Appendix_experiment_configurations}.

\paragraph{Pre-training Data.} Following prior work \citep{LayoutLM-v2}, we use IIT-CDIP Test Collection $1.0$ \citep{IIT-CDIP} containing $11$M document images for pre-training. Following DONUT \citep{DONUT}, we generate $2$M synthetic document images with text and layout annotations. Another four supplementary datasets with $0.4$M document images are also added to augment the diversity of pre-training data, including PubLayNet \cite{PubLayNet}, DocBank \citep{DocBank}, SciTSR \citep{SciTSR}, and IAM \citep{IAM}. We use our internal OCR tool to extract words with location coordinates from the IIT-CDIP and PubLayNet images. Words with locations are provided in IAM, SciTSR, and DocBank. Refer to Appendix \ref{Appendix_datasets} for more detailed data statistics. 

\paragraph{Evaluation Tasks.} We highlight that ViTLP are capable of handling both $1$) \textit{perception tasks} of document OCR and $2$) \textit{cognition tasks} of visual document understanding (VDU). To evaluate the comprehensive capabilities of ViTLP, we compare to baselines on each task as summarized in Table~\ref{tables:comparison}.

For OCR evaluation, we conduct two benchmark OCR sub-tasks, i.e., document text \textit{localization} and \textit{recognition}. We evaluate model performance on \textrm{SROIE} competition\footnote{\href{https://rrc.cvc.uab.es/?ch=13\&com=tasks}{https://rrc.cvc.uab.es/?ch=13\&com=tasks}} Task \#$1$ for text localization and Task \#$2$ for text recognition. The text localization task is evaluated by DetEval protocol \citep{DetEval} which calculates the precision, recall, and F$1$ based on the \textit{area of overlapping regions} between model predictions and ground-truth text coordinates. The text recognition task evaluates the word-level precision, recall, and F$1$ based on exact word match.

For VDU evaluation, we conduct three document understanding tasks. $1$) \underline{\textit{Form Understanding}}. Given a document image and its word entities, it is a sequential labeling task to predict the BIO tags for each textual entity. We use FUNSD \citep{FUNSD} which contains $199$ scanned forms, and the entities are labeled in four categories: \textit{Header}, \textit{Question}, \textit{Answer}, and \textit{Other}. FUNSD is divided into $149$ images for training and $50$ for testing. We report entity-level F$1$ as the evaluation score. $2$) \ul{\textit{Receipt Understanding}}. We use CORD \citep{CORD} containing $800$ training and $100$ testing images of real-world receipts. The receipt entities are labeled in $30$ categories. We use entity-level F$1$ for evaluation. $3$)  \ul{\textit{Document Classification}}. We conduct experiments on the RVL-CDIP dataset \citep{RVL-CDIP} containing $400$K scanned documents in $16$ classes. We adopt classification accuracy as the evaluation metric. For the sequence labeling tasks on FUNSD, we perform multi-segment fine-tuning on those samples whose entity-word sequences exceed the maximum decoder sequence length. This differs from previous work that truncates the input sequences into certain tokens, e.g., $512$ tokens in LayoutLMv$2$ \citep{LayoutLM-v2}.

Besides, we evaluate generative question answering tasks on the DocVQA \citep{DocVQA} and InfographicVQA \citep{InfoVQA} datasets. DocVQA consists of $12$K document images with $50$K QA pairs, and InfographicVQA contains $5.4$K document images with $30$K QA pairs. Since the answer word locations are not provided in the training sets, we use an OCR tool to locate the coordinates of answer words with heuristic text matching. In this way, we feed the answers with grounding coordinates to ViTLP for document VQA fine-tuning.

\subsection{OCR Evaluation Results}
We compare ViTLP with representative OCR baselines on SROIE $2019$ benchmark \citep{sroie2019}. The text localization baselines include CRAFT \citep{CRAFT}, YOLO-v$3$ \citep{YOLO-v3}, CTPN \citep{CTPN}, and EAST \citep{EAST}. The text recognition baselines include BiLSTM-ResNet, BiLSTM-CTC \citep{R2AM}, UNet-CRNN \citep{U-Net}, and TrOCR \citep{TrOCR}. Unlike conventional OCR models that first perform text localization and then use the localized text-regions for text recognition, ViTLP performs text localization and recognition in unified text-layout sequence generation, which does not need ground truth text-region inputs in the recognition task.

Table \ref{table:OCR} shows the OCR evaluation performance. ViTLP outperforms most baseline methods on both localization and recognition tasks. ViTLP underperforms TrOCR, given that TrOCR is a strong pre-trained model for two-stage OCR text recognition, while ViTLP performs text localization and recognition in one stage. Note that the SROIE training samples are few, i.e., only $626$ images, and the input text coordinates are at textline-level, which are different from our word-level pre-training input format and thus render it challenging to fine-tune our model. Nonetheless, ViTLP can still achieve competitive performance by fine-tuning on the limited samples without additional data augmentation \citep{TrOCR}, successfully adapting to output the textline coordinates that have never met in the pre-training phase. We also provide qualitative ViTLP zero-shot OCR examples in Appendix \ref{Appendix_OCR_examples}.
\begin{table}[t]
\centering
\resizebox{0.486\textwidth}{!}{
\begin{tabular}{lccc}
\toprule
& \multicolumn{3}{c}{\textit{Text Localization Task}} \\
\textbf{Method} & \textbf{Area-Precision} & \textbf{Area-Recall} & \textbf{Area-F1} \\
\midrule
CRAFT & 62.73 &	59.94 &	61.31 \\
YOLO-v$3$ & 77.29 & 79.32 & 78.29 \\
CTPN & 81.14 & 87.23 & 84.07 \\
EAST & 85.07 & 87.17 & 86.11 \\
ViTLP & 91.62 & 91.68 & 91.65 \\
\hline\hline
& \multicolumn{3}{c}{\textit{Text Recognition Task}} \\
\textbf{Method} & \textbf{Word-Precision} & \textbf{Word-Recall} & \textbf{Word-F1} \\
\midrule
BiLSTM-ResNet & 74.05 & 77.81 & 75.88 \\
BiLSTM-CTC & 83.38 & 87.37 & 85.33 \\
UNet-CRNN & 85.77 & 86.48 & 86.12 \\
TrOCR$^\dagger$ & 95.89 & 95.74 & 95.82 \\
ViTLP & 93.07 & 92.52 & 92.79 \\
\bottomrule
\end{tabular}
}
\caption{OCR text localization and recognition results on SROIE $2019$ benchmark. $^\dagger$TrOCR uses the ground-truth cropped image regions as inputs, whereas ViTLP performs text localization and recognition in a unified stage. All scores are reported in percentage.}
\label{table:OCR}
\end{table}

\subsection{VDU Evaluation Results}
We compare ViTLP with competitive pre-trained baselines including \romannumeral 1) general method RoBERTa \citep{Roberta}, \romannumeral 2) discriminative VDU models: LayoutLM \citep{LayoutLM-v1}, SPADE \citep{SPADE}, SelfDoc \citep{SelfDoc}, TITL \citep{TILT}, LayoutLMv$2$ \citep{LayoutLM-v2}, LiLT \citep{LILT}, FormNet \citep{FormNet} and \romannumeral 3) generative VDU model DONUT \citep{DONUT}. Table \ref{table:VDU} shows the VDU task performance.
\begin{table*}
\centering
\resizebox{\textwidth}{!}{
\begin{tabular}{lcccccc}
\toprule 
\textbf{\tabincell{l}{Method}} & \textbf{\tabincell{c}{Modeling Type}} & \textbf{\tabincell{c}{\# Param.}} & \textbf{\tabincell{c}{Maximum \\ Doc-Length}} & \textbf{\tabincell{c}{FUNSD (F1)}} &\textbf{\tabincell{c}{CORD (F1)}} & \textbf{\tabincell{c}{RVL-CDIP (Acc)}} \\
\midrule
RoBERTa\textsubscript{\textrm{base}}~\cite{Roberta} & & 125M & 512 & 66.48 & 93.54 & 90.06 \\
LayoutLM\textsubscript{\textrm{base}}~\cite{LayoutLM-v1} & & 160M & 512 &79.27 & \textendash & 94.42 \\
SPADE~\cite{SPADE} & & 110M & 512 & 70.50 & 91.50 & \textendash \\
SelfDoc~\cite{SelfDoc} & Discriminative & 137M & 1024 & 83.36 & \textendash & 93.81 \\
TILT\textsubscript{\textrm{base}}~\cite{TILT} & (w/ OCR Input) & 230M & 512 & \textendash & 95.11 & 95.25 \\
LayoutLMv2\textsubscript{\textrm{base}}~\cite{LayoutLM-v2} & & 200M & 512 & 82.76 & 94.95 & 95.25 \\
LiLT\textsubscript{\textrm{base}}~\cite{LILT} & & \textendash & 512 & 88.41 & 96.07 & 95.68 \\
FormNet~\cite{FormNet} & & 217/345M$^\dagger$ & 1024 & 84.69 & 97.28 & \textendash \\
\midrule
\textsc{DONUT}~\cite{DONUT} & Generative & 259M & 1536 & \textendash & 84.10 & 95.30 \\
ViTLP & (w/o OCR Input) & 253M & Any-length & 87.61 & 95.59 & 95.36 \\
\bottomrule
\end{tabular}}
\caption{VDU evaluation results on form understanding (FUNSD), receipt understanding (CORD), and document classification (RVL-CDIP).  $^\dagger$ FormNet has different sizes of $217$M and $345$M for FUNSD and CORD \citep{FormNet}. ``Maximum Doc-Length'' denotes the maximum tokens of an input text sequence that the model can handle.}
\label{table:VDU}
\end{table*}
\paragraph{Information Extraction.} According to Table \ref{table:VDU}, our model achieves better F$1$ scores compared to most baselines on FUNSD and CORD. The results indicate that ViTLP can develop a thorough understanding of form/receipt structures from images. Nonetheless, ViTLP underperforms the best discriminative baselines, i.e., LiLT on FUNSD and FormNet on CORD. We believe this is because pre-trained discriminative VDU models have natural advantages over generative models for the information extraction task, which is formulated as token-level classification. Besides, ViTLP outperforms DONUT, proving that layout modeling is as necessary as language modeling to generative VDU models. For example, for the CORD images, entities with the same semantic label \texttt{<menu.price>} are always located in the same rightmost column of the receipt, sharing adjacent layout coordinates. Layout modeling can help generative VDU models better extract such structural-aware information.

\paragraph{Document Classification.} From Table~\ref{table:VDU}, we can see that ViTLP achieves the second best performance on classification accuracy. We also observe that the performance among TILT, LayoutLMv$2$, DONUT, and ViTLP are quite close. This may be because document classification is a coarse-grained task, wherein the vision modality contributes the most to classification performance, and the OCR text modality brings an incremental gain. Though ViTLP is suboptimal compared to LiLT, OCR-free generative methods are more flexible and lightweight because no pre-processed OCR texts are needed for input.

\subsection{Further Discussion}
\subsubsection{Ablation Study}
We conduct ablation studies on the effect of hierarchical text-layout modeling and multi-segment pre-training scheme. We compare ViTLP with three variants: \romannumeral 1) pre-training with the language modeling objective only, without the layout modeling objective; \romannumeral 2) truncating long input document sequences in pre-training, without the multi-segment strategy; \romannumeral 3) generating four layout coordinate tokens for each word in a long flatten sequence, without hierarchical text-layout modeling.

\begin{table}[t]
\centering
\resizebox{0.486\textwidth}{!}{
\begin{tabular}{lcc}
\toprule
\textbf{Ablation Variants} & \textbf{FUNSD (F1)} & \textbf{CORD (F1)} \\
\midrule
ViTLP & 87.61 & 95.59 \\ \hline
w/o layout modeling & 81.42 & 91.54 \\
w/o multi-segment training & 86.73 & 95.01 \\
w/o hierarchical modeling & 86.28 & 94.86 \\
\bottomrule
\end{tabular}
}
\caption{Ablation model performance on the information extraction tasks.}
\label{table:ablation_study}
\end{table}
Table \ref{table:ablation_study} displays the ablation performance. We can observe that discarding the layout modeling objective leads to a substantial performance drop, i.e., $6.19$ and $4.05$ F$1$ drops on FUNSD and CORD. The results suggest that generative pre-training on the layout modality can enhance the document understanding capability of VDU models. Besides, truncating long document inputs without the multi-segment pre-training strategy leads to lower performance. We believe that the multi-segment pre-training scheme enables ViTLP to model complete text and layout tokens of the pre-training corpora, which benefits the pre-trained model performance. We can also see that removing hierarchical text-layout modeling causes performance descent. It validates that hierarchical modeling is effective for interleaved text-layout information fusion.

\subsubsection{Generative Document VQA}
\begin{table}[t]
\centering
\resizebox{0.486\textwidth}{!}{
\begin{tabular}{lcc}
\toprule
\textbf{Generative Model} & \textbf{DocVQA} & \textbf{InfographicVQA} \\
\midrule
Dessurt~\cite{Dessurt} & 63.2 & \textendash \\
DONUT~\citep{DONUT} & 67.5 & 11.6 \\
ViTLP & 65.9 & 28.7 \\
\bottomrule
\end{tabular}
}
\caption{The results are reported on Average Normalized Levenshtein Similarity (ANLS) between the model-generated answers and ground truth.}
\label{table:DocVQA}
\end{table}
\begin{figure*}[t]
\centering
\includegraphics[width=162.5mm]{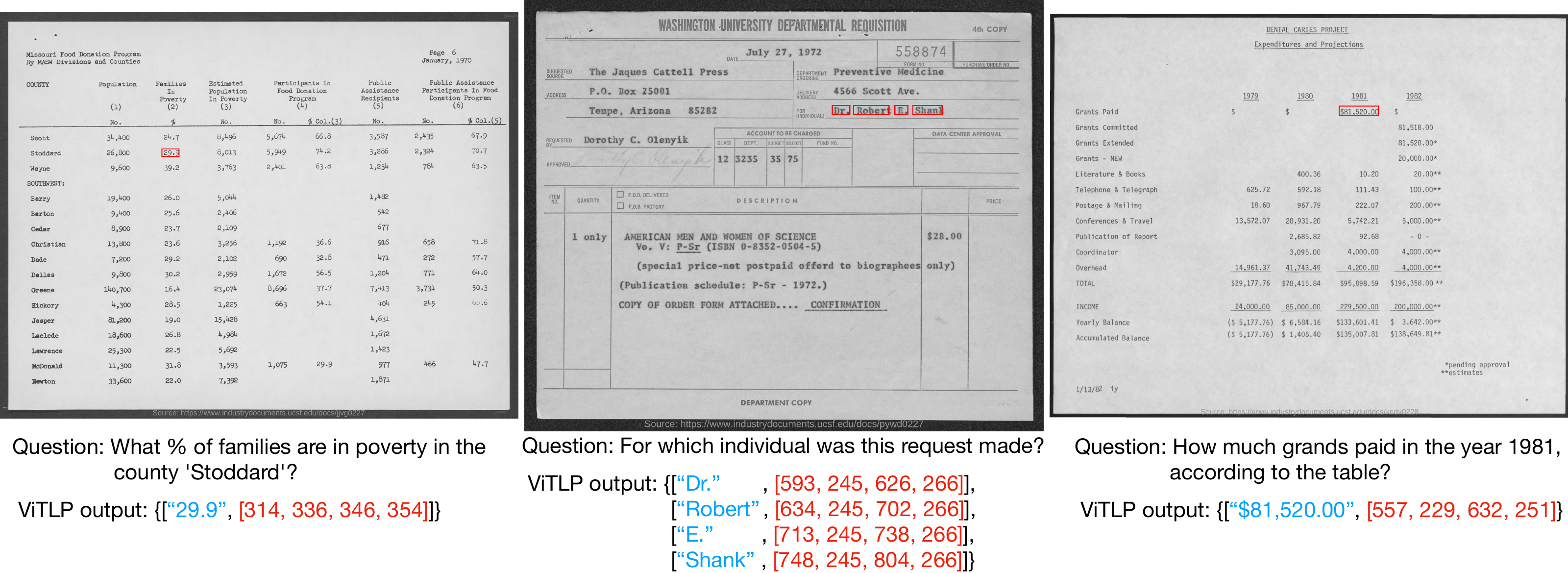}
\captionsetup{font=10pt}
\caption{Visualization of ViTLP generated answers on DocVQA. The ViTLP output answer sequences consist of answer words ({\color{blue}in blue}) and corresponding location coordinates ({\color{red}in red}). For direct visualization, we draw the region of interest (ROI) \textbf{referring to the output layout coordinates} on the image.}
\label{Figures:visualization}
\end{figure*}
\paragraph{Results and Analysis.} Table \ref{table:DocVQA} presents the performance of generative VDU models on DocVQA and InfographicVQA datasets. We can see that ViTLP underperforms DONUT by a slight margin on DocVQA and surpasses DONUT by a significant margin on InfographicVQA. As discussed in \citet{DONUT}, DocVQA images are similar to the pre-training IIT-CDIP images, pre-training data quality may have a considerable influence on the performance of DocVQA. The average results show that ViTLP develops better overall document VQA performance than the strong generative model DONUT, which validates the effectiveness of our generative pre-training approach.

\paragraph{Document VQA with Interpretable Grounding.} Owing to the fine-grained \textit{word-level} grounding capability learned in the pre-training stage, ViTLP can be fine-tuned to predict the regions of interest (ROI) associated with the generated answers, which is unprecedented to prior document VQA models. As shown in Figure \ref{Figures:visualization}, the output ROI grounding-boxes \textit{as visual rationales} can help humans easily verify the model-generated answers, making the answer generation process \textit{interpretable to humans} where the model output derives from. See more examples of grounding document VQA with ViTLP in Appendix \ref{Appendix_VQA_examples}.

\section{Related Work}
Visual document processing with multimodal pre-training is widely studied. From the perspectives of the document processing pipelines and model architectures, existing works can be generally divided into strands of research as listed below.

\textbf{OCR-based Methods.} Most initial VDU efforts adopt OCR tools to localize and recognize document layouts and texts, and then feed them to the multimodal pre-trained models~\citep{LayoutLM-v2, DocFormer, structurallm, ErnieLayout, SelfDoc, WuKongReader, Formnet-v2}. These methods usually involve multiple pre-training objectives over the vision, text, and layout. For instance, document text location~\cite{LayoutLM-v1, LayoutLM-v2}, paragraph and table regions~\citep{SelfDoc, MGDoc} are rich in structural information to align visual features with text embeddings. Though promising, these pipeline models suffer from heavy OCR pre-processing overhead. Moreover, incorrect OCR results may propagate errors to downstream tasks like document question answering~\citep{DONUT}.

\textbf{OCR-free Methods.} There appear recent studies \citep{DONUT, Pix2Struct, PreSTU} that jointly consider text reading and understanding without external OCR pipelines. For instance, \citet{DONUT} takes document images as input to the model without prerequisite OCR results and conducts visual language pre-training. \citet{Pix2Struct} further improves the pre-training objectives over large-scaled visual webpage corpora. \citet{PreSTU} employs multiple pre-training tasks jointly to encourage the pre-trained model to learn text recognition capability explicitly and spatial reasoning capability implicitly.

Our research falls within the OCR-free branch. Different from existing works, we first study generative joint text-layout modeling conditioned on input document images. Our empirical results also validate that layout information not only enhances the learned representations for downstream VDU tasks but also can make the generation outputs more interpretable with visual groundings.

\textbf{LLM-backbone Methods.} Most recent studies leverage large language models (LLMs) to tackle multimodal document tasks \citep{LLaVAR, mPLUG-DocOwl, DocLLM}. LLaVAR \citep{LLaVAR} inherits LLaVA architecture \citep{LLaVA} which directly projects the visual features to LLM embeddings and performs instruction tuning on visual document data. DocLLM \citep{DocLLM} uses spatial attention to inject $2$D layout information into Llama $2$ \citep{LLaMA2} with supervised fine-tuning and first enables LLMs to process document information extraction tasks. Thanks to LLMs' powerful reasoning and generation abilities, utilizing LLMs for visual document processing has become a prominent research trend.
\section{Conclusion}
We propose visually guided generative text-layout pre-training (ViTLP) to enhance visual document processing covering the OCR and VDU tasks. In the pre-training phase, ViTLP optimizes hierarchical language and layout modeling objectives to generate interleaved text-layout target sequences. Moreover, the proposed multi-segment pre-training scheme enables ViTLP to process long documents with arbitrary lengths. ViTLP can function as a native OCR model to locate and recognize texts of document images. Experiments also show that ViTLP achieves superior performance on various VDU tasks with document grounding capability.

\section*{Limitations}
Our community has entered the era of large language models with multimodal capabilities \citep{InstructBLIP, GPT4-v}. However, regarding the model size, ViTLP is still a rather small-scale pre-trained model\footnote{It is because we commenced the ViTLP project in mid-2022 and finished pre-training in early 2023, see the first version at \href{https://openreview.net/forum?id=ARtBIBAmNR}{https://openreview.net/forum?id=ARtBIBAmNR}.}, which limits its potential to become an interactive and generalized document AI assistant. In future work, we plan to explore two paths: \romannumeral 1) scaling up ViTLP with more parameters and training data, extending it to a more powerful foundation document model; \romannumeral 2)~integrating ViTLP's \textit{document-specific} text-layout image encoder with \textit{generalized} advanced LLMs \citep{Vicuna, LLaMA2} and visual instruction tuning \citep{LLaVA, MiniGPT-4} to build up an interactive document AI assistant.

\textbf{Remarks and Future direction.} \romannumeral 1) ViTLP processes document images already calibrated in angle. Hence, we use $4$ coordinates to represent the localization of words. It is feasible to pre-train ViTLP to generate $8$ coordinates which can represent the angle of words. We choose word-level segmentation for pre-training because \textit{a word is the elementary unit of document texts}. Word-level segmentation is also beneficial to fine-grained grounding, e.g., VQA with answer-word grounding. \romannumeral 2) We propose a multi-segment processing scheme to permit long sequence lengths on the \textit{decoder side}. However, the document pixel inputs are also constrained by the resolution on the ViT \textit{encoder side}. For the problem of long document processing, ViTLP only tackles the half. Processing document images with high resolutions and multiple pages is an intriguing problem for future research.

\section*{Acknowledgements}
We appreciate constructive comments from anonymous ARR reviewers. We thank Bin Liang from CUHK for valuable discussion. This research work is partially supported by CUHK direct grant No. $4055209$ and CUHK Knowledge Transfer Project Fund No. KPF$23$GWP$20$.

\bibliography{anthology, custom}
\clearpage
\appendix
\begin{table}[t]
\centering
\resizebox{0.5\textwidth}{!}{
\begin{tabular}{cccc}
\toprule
Dataset & Size & Proportion & Document Type \\
\midrule
IIT-CDIP & $10,816,672$ & $81.89\%$ & Scanned Document \\
SynthDog & $2,000,000$ & $15.14\%$ & Synthetic Document \\
PublayNet & $261,076$ & $1.98\%$ & Scientific Paper \\
DocBank & $125,815$ & $0.95\%$ & Arxiv Paper \\
SciTSR & $3,536$ & $0.03\%$ & Figure and Table \\
IAM & $1,198$ & $0.01\%$ & Hand Written \\
\bottomrule
\end{tabular}}
\caption{Pre-training dataset statistics.}
\label{dataset_statitics}
\end{table}
\section{Experiment Details}
\subsection{Pre-training Data Statistics}\label{Appendix_datasets}
Table \ref{dataset_statitics} shows the pre-training data statistics. Following previous work, e.g., LayoutLMv$2$ \citep{LayoutLM-v2}, we use $11$M IIT-CDIP document images as the main pre-training data. Besides, we follow \citet{DONUT} and \citet{Dessurt} to include $2$M machine-rendered synthetic documents for generative pre-training. Specifically, we adapt the official SynthDog generator\footnote{\href{https://github.com/clovaai/donut/tree/master/synthdog}{https://github.com/clovaai/donut/tree/master/synthdog}} to generate synthetic document images with text and layout metadata. The other four corpora, i.e., PublayNet, DocBank, SciTSR, and IAM, account for only $\sim3\%$ pre-training data whereby we aim to improve the diversity of pre-training document types.

\begin{figure}
\centering
\includegraphics[width=1.04\linewidth]{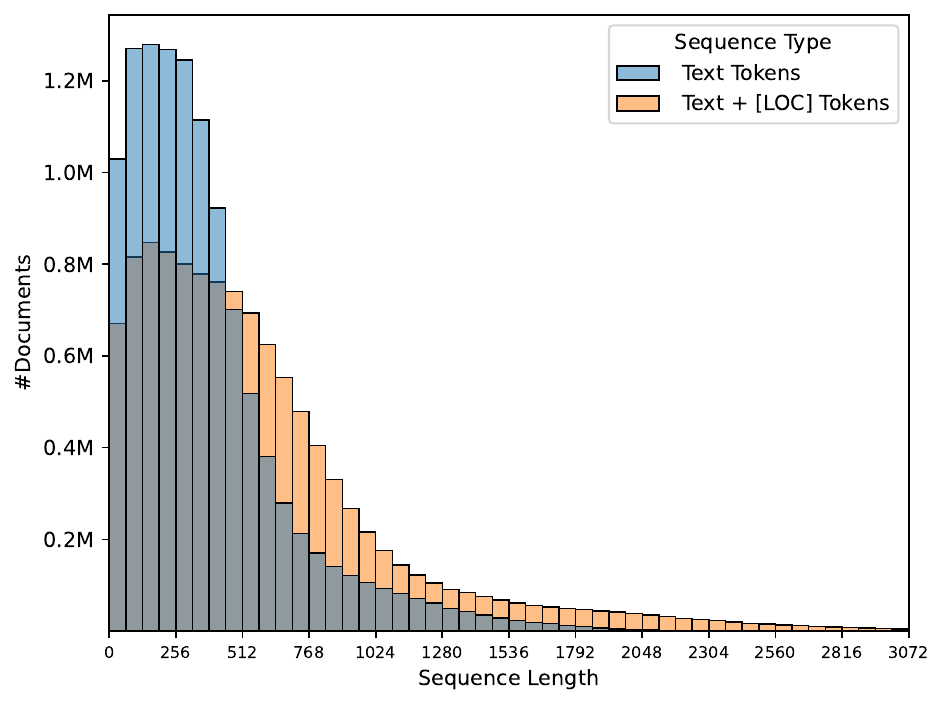}
\caption{Distribution of document sequence lengths. The text sequences are tokenized by the standard BPE tokenizer \citep{GPT2}. }
\label{document_length}
\end{figure}

The distribution of document sequence lengths is displayed in Figure \ref{document_length}. The number of text-layout sequence tokens follows a \textit{long-tailed distribution}: there exist some long documents with the sequence lengths ranging from $1024$ to $3072$. This brings a trade-off to pre-training. With a relatively short sequence length (e.g., $512$ tokens in LayoutLMv$2$), language modeling on long documents is incomplete, as the sequence tokens are truncated and wasted. However, with a relatively long sequence length (e.g., $3072$), the GPU computation and memory overload would become prohibitive, which also forbids large batch sizes for better performance.\footnote{Even assuming sufficient computation resources, the long-tailed distribution of document lengths would also cause enormous padding tokens in long sequence input to Transformers, leading to considerable waste of computational resources.} The proposed multi-segment pre-training scheme can circumvent this bitter trade-off. Notably, the multi-segment processing scheme can be directly applied to long document fine-tuning (and inference). For example in the OCR and sequence labeling tasks, ViTLP also employs the multi-segment scheme to process the long documents by multiple segments with prefix context tokens.

\subsection{Fine-tuning Hyperparameter Settings}\label{Appendix_experiment_configurations}
\paragraph{OCR Text Localization and Recognition.} Fine-tuning ViTLP for text localization and recognition follows the same objective Eq.~\eqref{eq:final_obj} as pre-training. Since the SROIE $2019$ \citep{sroie2019} training set is rather small containing only $626$ images, we fine-tune ViTLP for $10$ epochs with the batch size of $1$. The used learning rate and weight decay are $2e$-$5$ and $1e$-$2$. The input image resolution remains the same as pre-training, i.e., $1920\mathrm{\times}1600$.

\paragraph{Information Extraction.} For FUNSD \citep{FUNSD}, the selected learning rate and weight decay are $1e$-$4$ and $1e$-$2$. For CORD \citep{CORD}, the selected learning rate and weight decay\footnote{For CORD, we search the configuration of learning rate in $\{2e$-$4,1e$-$4,5e$-$5,3e$-$5,2e$-$5,1e$-$5\}$ and weight decay in $\{1e$-$2,1e$-$4\}$.} are $5e$-$5$ and $1e$-$4$. For both datasets, we fine-tune ViTLP for $75$ epochs with the batch size of $8$, using the same input image resolution as pre-training. Following the practice of prior work \citep{LayoutLM-v3, Formnet-v2}, we use the shared segment-level layout coordinates as input instead of word-level coordinates, which can benefit the token classification accuracy in sequence labeling.

\paragraph{Document Classification.} We use the learning rate of $1e$-$4$ and weight decay of $1e$-$2$ for the document classification task. We fine-tune ViTLP for $100$ epochs with the global batch size of $320$. The input image resolution is the same as pre-training.

\paragraph{Document VQA.} Since the layout coordinates of answer words are not provided in the DocVQA \citep{DocVQA} and InfographicVQA \citep{InfoVQA} datasets, we first conduct OCR on the training document images to obtain the texts with bounding-box coordinates. Then we apply a heuristic text-matching method to assign corresponding bounding-box coordinates to the answer words. It is worth noting that for the "Yes/No" questions that have no grounding answers on the images, we train ViTLP to generate a special answer token \texttt{[ANS\_YES]} or \texttt{[ANS\_NO]} without layout coordinates. For both datasets, we fine-tune ViTLP for $60$ epochs with a batch size of $128$. We use a learning rate of $3e$-$5$. Since the document images are high-resolution, for DocVQA, we set the fine-tuning image resolution as $2304\mathrm{\times}1920$ which is multiplied by $1.2$ based on the pre-training resolution. For InfographicVQA, the fine-tuning image resolution is set as $3200\mathrm{\times}1600$. From our empirical experiments, we find that input image resolution is essential to document VQA performance, especially for InfographicVQA.

\section{Implementation Details of Sequential Layout Head}\label{Appendix_RNN_layout_head}
Given that multimodal interaction is learned by the stacked Transformer text-layout decoder layers, the LM and layout heads hereby function as a prober to output the next word and coordinate predictions. As introduced in Sec \ref{local_layout_modeling}, the layout head predicts output probability $\mathrm{Prob}(\m L_{i,j})$ of the four coordinates $\{\m L_{i,j}\}_{j=1}^4 = \{z_{x_1}, z_{y_1}, z_{x_2}, z_{y_2}\}_i$ based on the $i$-th global \texttt{[LOC]} token's final hidden state $\m H_{i,0}=\m H_i^{VTL}\in\mathbb{R}^{d}$ as follows.
\begin{equation*}
\begin{split}
\begin{cases}
\m H_{i,1}=\mathrm{GELU}\big(\m W_{h}\m H_{i,0}\big) \\
\m H_{i,2}=\mathrm{GELU}\big(\m W_{h}\m H_{i,1} + \mathrm{\m E}^{'}_{x}(\m L_{i,1})\big) \\
\m H_{i,3}=\mathrm{GELU}\big(\m W_{h}\m H_{i,2} + \mathrm{\m E}^{'}_{y}(\m L_{i,2})\big) \\
\m H_{i,4}=\mathrm{GELU}\big(\m W_{h}\m H_{i,3} + \mathrm{\m E}^{'}_{x}(\m L_{i,3})\big)\\
\end{cases} \\
\end{split}
\end{equation*}
\begin{equation*}
\mathrm{Prob}(\m L_{i,j})=\textrm{Softmax}\big(\m W_{L}\m H_{i,j}\big), \;\;j\in\{1,2,3,4\}
\end{equation*}
The coordinate tokens are quantized into a discrete range of $[0, 1000]$, making the layout-token vocabulary size of $|L|=1001$. The layout head's parameters are lightweight including a hidden matrix $\m W_{h}\in\mathbb{R}^{d\times d}$, two embeddings 
$\mathrm{\m E}^{'}_{x}(\cdot)\in\mathbb{R}^{d}$ and $\mathrm{\m E}^{'}_{y}(\cdot)\in\mathbb{R}^{d}$, and a linear projection $\m W_{L}\in\mathbb{R}^{|L|\times d}$. We use the same GELU activation \citep{GELU} as in the Transformer layers. The layout head works sequentially, which is similar to a vanilla RNN, as each coordinate decoding step also considers the information of previous coordinates. Compared with naively using four independent linear heads, the sequential layout head can capture the spatial relation among the output coordinates (e.g., $x_{1} < x_{2}$ and $y_{1} < y_{2}$), bootstrapping more accurate coordinate prediction.

\begin{figure*}[ht]
\hskip-13mm
\includegraphics[width=186mm]{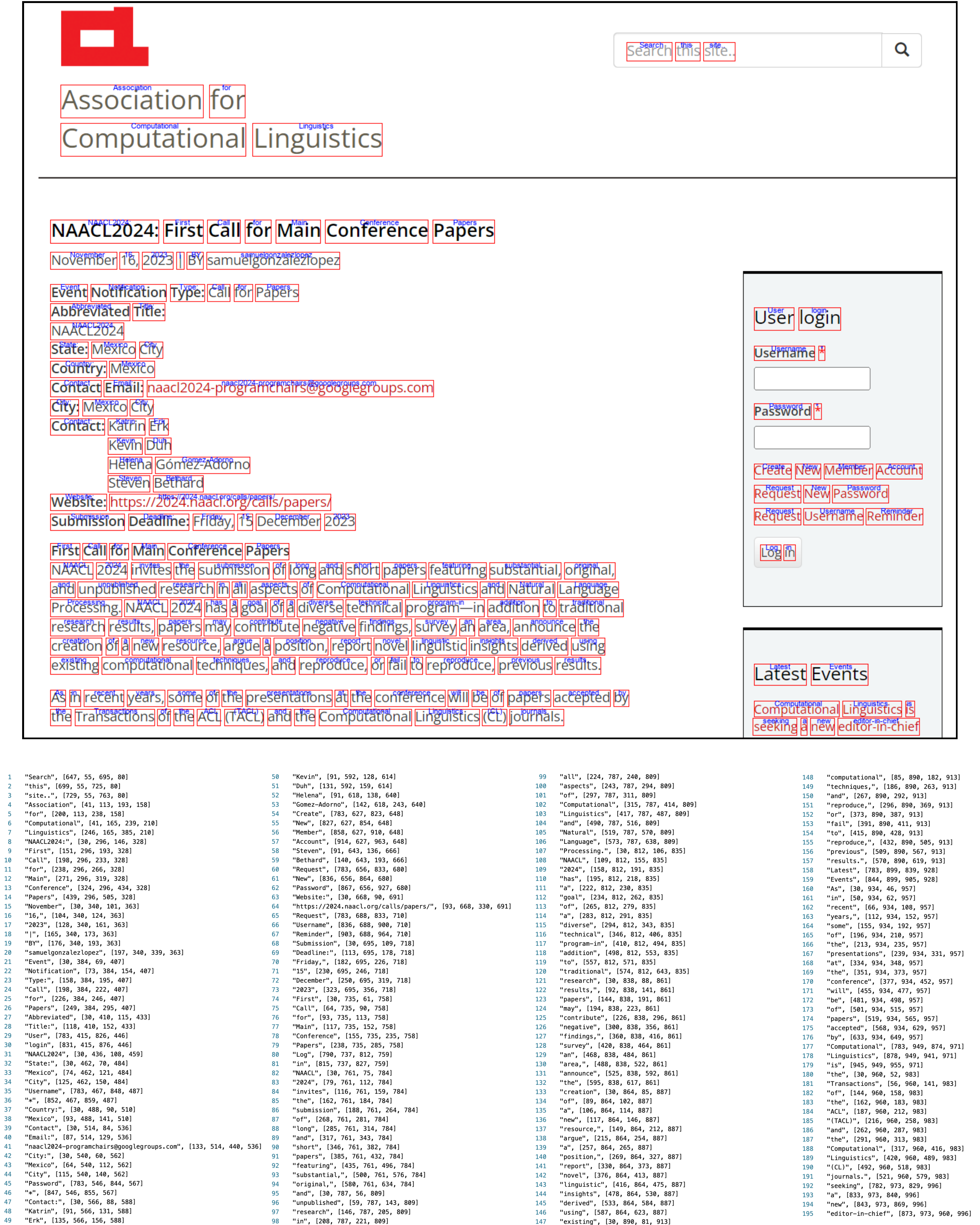}
\captionsetup{font=10pt}
\caption{ViTLP OCR results on a webpage. For comprehensive visualization, we render the output texts (in blue) and bounding boxes (in red) according to the ViTLP's interleaved output sequence.}
\vskip4mm
\label{Figures:ocr-2}
\end{figure*}
\begin{figure*}[ht]
\vskip-11mm
\hskip-5.5mm
\frame{\includegraphics[width=170.5mm]{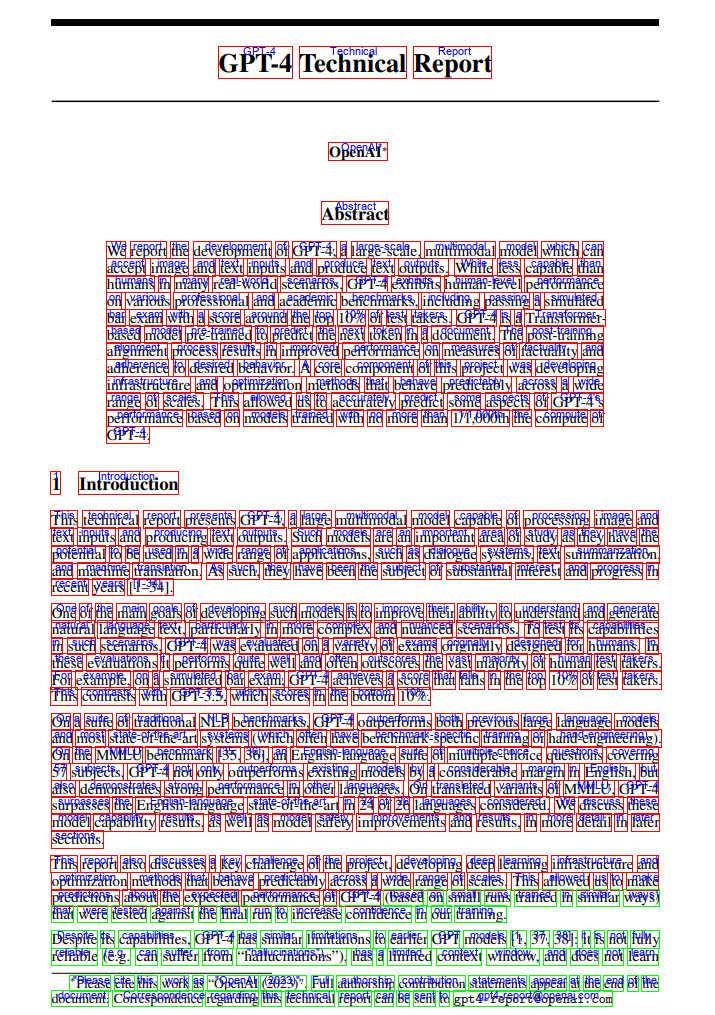}}
\vskip-1mm
\captionsetup{font=10pt}
\caption{ViTLP OCR results on a paper. For comprehensive visualization, we render the words and bounding boxes according to ViTLP's interleaved output sequence. The shown generated OCR results comprise two segments, as the generated tokens reach the decoder sequence length ($M=1024$) in the first segment generation, and the generation process continues by the second segment. The bounding boxes of the first segment are in red, and the second are in green.}
\label{Figures:ocr-1-vis}
\end{figure*}

\begin{figure*}[ht]
\vskip-17.5mm
\hskip-12mm
\includegraphics[width=184mm]{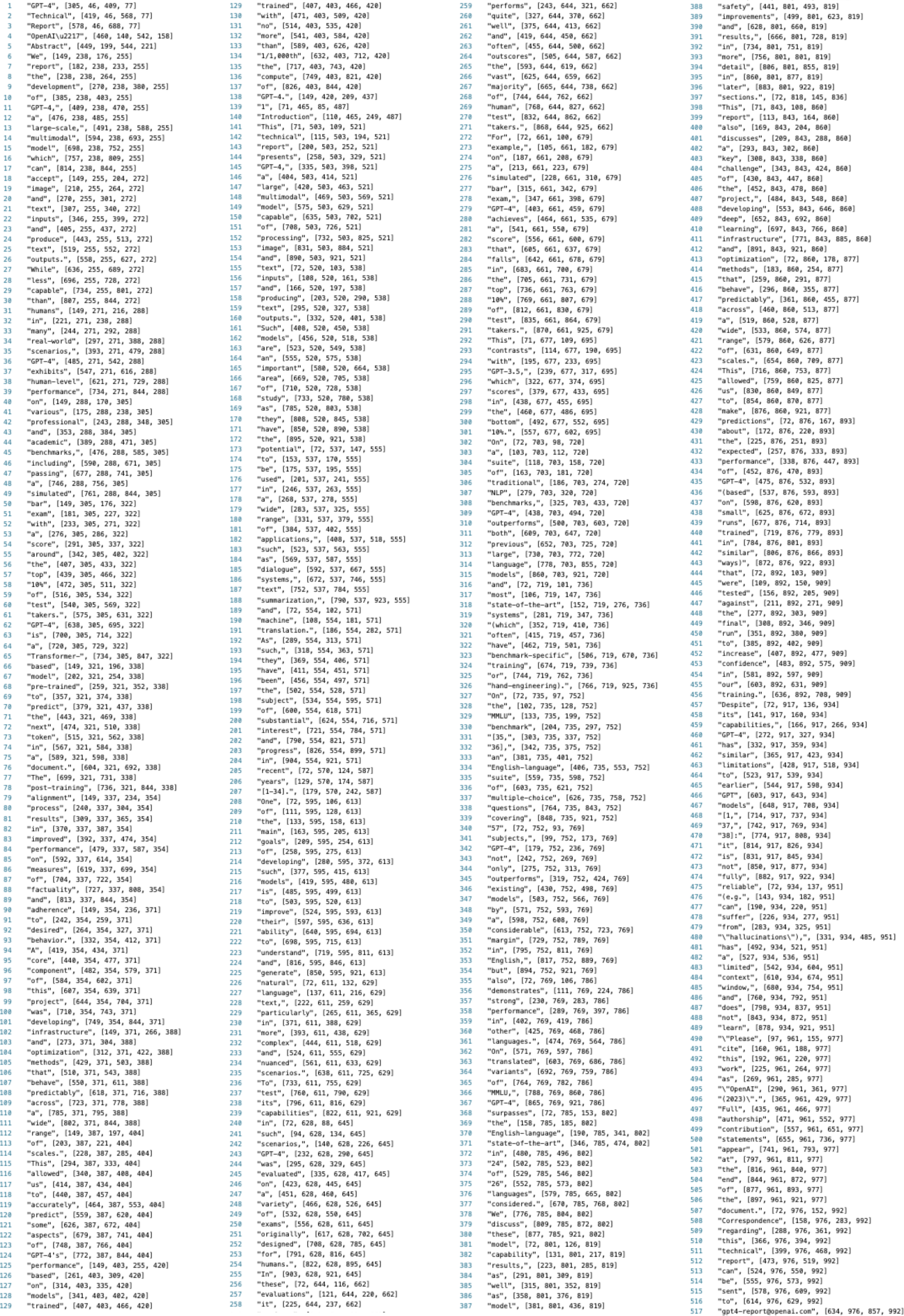}
\vskip-2.75mm
\captionsetup{font=10pt}
\caption{ViTLP OCR results as visualized in Figure \ref{Figures:ocr-1-vis} above.}
\label{Figures:ocr-1-text}
\end{figure*}

\section{Qualitative Cases of ViTLP Document OCR Functionality}\label{Appendix_OCR_examples}
Figure \ref{Figures:ocr-2} to \ref{Figures:ocr-1-text} demonstrate ViTLP's functionality on zero-shot document OCR. ViTLP outputs the interleaved OCR sequence consisting of words and corresponding bounding boxes.

\begin{figure*}[ht]
\includegraphics[width=162.5mm]{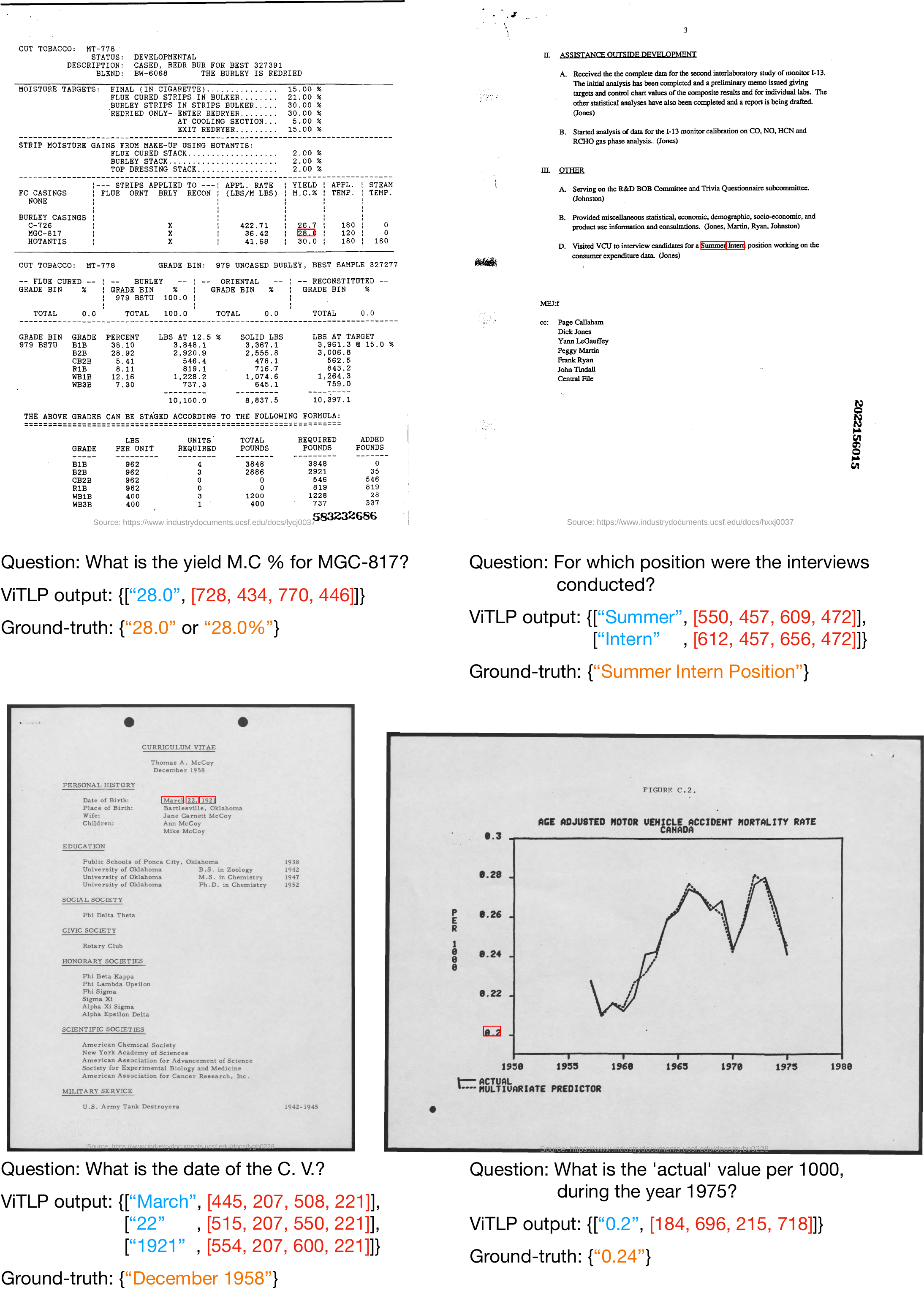}
\captionsetup{font=10pt}
\caption{Four examples (two successful cases \& two failure cases) of ViTLP document VQA outputs with grounding locations.}
\label{Figures:vqa}
\end{figure*}
\section{Qualitative Cases of ViTLP Document VQA with Grounding Capability}\label{Appendix_VQA_examples}
Figure \ref{Figures:vqa} showcases the ViTLP's VQA outputs on DocVQA with grounding capability. The top two examples are successful cases, and the bottom two are failure cases.
\end{document}